# Using Convolutional Neural Networks for Determining Reticulocyte Percentage in Cats

Krunoslav Vinicki[1]   Pierluigi Ferrari[2]   Maja Belić[1]   Romana Turk[1]

## Abstract

Recent advances in artificial intelligence (AI), specifically in computer vision (CV) and deep learning (DL), have created opportunities for novel systems in many fields. In the last few years, deep learning applications have demonstrated impressive results not only in fields such as autonomous driving and robotics, but also in the field of medicine, where they have, in some cases, even exceeded human-level performance. However, despite the huge potential, adoption of deep learning-based methods is still slow in many areas, especially in veterinary medicine, where we haven't been able to find any research papers using modern convolutional neural networks (CNNs) in medical image processing. We believe that using deep learning-based medical imaging can enable more accurate, faster and less expensive diagnoses in veterinary medicine. In order to do so, however, these methods have to be accessible to everyone in this field, not just to computer scientists. To show the potential of this technology, we present results on a real-world task in veterinary medicine that is usually done manually: feline reticulocyte percentage. Using an open source Keras implementation of the Single-Shot MultiBox Detector (SSD) model architecture and training it on only 800 labeled images, we achieve an accuracy of 98.7% at predicting the correct number of aggregate reticulocytes in microscope images of cat blood smears. The main motivation behind this paper is to show not only that deep learning can approach or even exceed human-level performance on a task like this, but also that anyone in the field can implement it, even without a background in computer science.

## 1. Introduction

Convolutional Neural Networks (CNN) are a class of deep artificial neural networks that is loosely inspired by the structure and function of the mammalian visual cortex [1]. CNNs show similar levels of abstraction to our visual cortex: Earlier layers of the neural network detect simple features like edges, while later layers detect gradually more complex shapes and objects [2, 3]. These multiple levels of abstraction allow CNNs to learn to extract complex features directly from raw input images [3].

[1] University of Zagreb, Faculty of Veterinary Medicine
[2] Technical University of Berlin, Department of Mathematics
Correspondence to: Krunoslav Vinicki <kvinicki@gmail.com>, Pierluigi Ferrari <pierluigi.ferrari@gmx.com>.

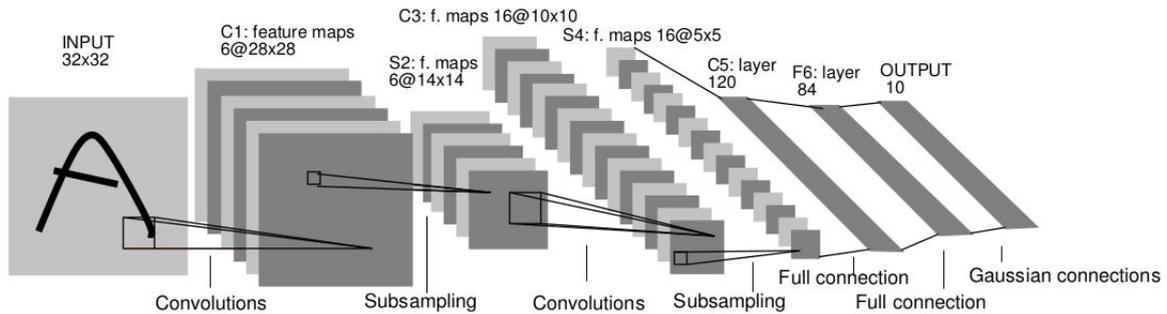

Fig. 1. Convolutional neural network LeNet-5 from LeCun et. al. [4].

The neurons of a CNN are arranged in layers (Fig. 1). Four main types of layers are most commonly used to construct CNNs: convolutional layers, pooling layers, batch normalization layers, and fully connected layers. The output of a convolutional layer is computed by sliding a filter (or "kernel") across the spatial dimensions of the input and computing the dot product between the filter and the input at the respective position, thereby computing a convolution or cross-correlation over the input. This implies that (1) the kernel of a given convolutional layer is shared across all spatial locations of its input, taking advantage of the fact that a given pattern in the input represents the same feature regardless of where it is located, and (2) each neuron in a convolutional layer is only affected by a specific local region of the input image, called that neuron's "receptive field". Pooling layers sub-sample their input spatially and don't have trainable parameters. Their purpose is two-fold: By down-sampling the spatial resolution of the input, they increase the translation and distortion invariance of objects in the input and reduce the computational complexity for subsequent layers. Stacking multiple convolutional and pooling layers leads to increasingly more abstract features and increasingly larger receptive fields in the later layers of the network [4]. Many successful network architectures used for image classification add fully connected layers at the top of the network that serve as a classifier that uses the features extracted by the underlying CNN. Many more recent architectures, however, so-called fully convolutional networks (FCN), get rid of fully connected layers entirely and instead rely on convolutional layers as the only type of layers with trainable parameters. We won't go into any detail about batch normalization layers apart from mentioning that they have a stabilizing effect on the training process that speeds up the learning process and usually leads to better overall training results.

Even though CNNs were first introduced almost three decades ago [4], their extremely successful use to solve practical task has only began very recently. In the past few years, deep learning has also increasingly been applied to problems in medicine. This has been due to the availability of much larger labeled image datasets, advances in hardware (especially GPUs), the development of better network architectures and learning algorithms, new open source libraries, and the availability of pre-trained neural networks [5, 6]. Recently, CNNs have been used to solve a number of problems in medical imaging, yielding an improvement over previous state-of-the-art results [7]. In some cases, CNNs even outperformed humans in this tasks [8, 9, 10]. Unfortunately, despite the successful application of deep learning in many areas, we are not aware of any applications in veterinary imagery. We argue that this needs to change though, because not only can deep learning enable more accurate, faster and less expensive diagnoses in veterinary medicine, but it can also be implemented by veterinarians without a background in computer science. Problems can only be solved by people who are familiar with them. Computer scientists won't solve the majority of problems in veterinary medicine, but veterinarians can, not just because they are the ones who are confronted with those problems every day, but also because they have the medical expertise needed for collecting and labeling medical images.

In order to show the potential of deep learning in veterinary medicine we decided to apply it in determining feline reticulocyte percentage. Reticulocytes are immature erythrocytes and, being released from bone marrow in increased numbers as a response to elevated erythropoietin levels, they are an important indicator of bone marrow erythropoiesis. In other words, they allow assessment of whether a cat's bone marrow, given sufficient time, is responding to an anemia by increasing erythrocyte production [11]. For this reason they are used to classify anemias into regenerative and nonregenerative categories. Since reticulocytes are still synthesizing hemoglobin, they can be recognized by the presence of RNA in their cytoplasm after staining with supravital dyes like new methylene blue (NMB). With these dyes we can recognize two types of reticulocytes in cats: more immature aggregate reticulocytes with big aggregates of RNA and more mature punctate reticulocytes with one or more small granules of residual RNA [12]. It is important to note that aggregate reticulocytes persist for 12-24 hours in circulation, after which they mature into punctate reticulocytes that can persist for another 7-10 days [11]. Therefore, only aggregate reticulocytes reflect the current bone marrow response to anemia and are included in reticulocyte count.

Reticulocyte percentage is not only a generally important problem in veterinary medicine, but also a very interesting one from the standpoint of deep learning. It is a task that is usually done manually and, although automated systems exist, manual counting by light microscopy still remains the gold standard [12]. This is because an automated system like flow cytometry can show falsely elevated aggregate reticulocytes in presence of some artifacts and erythrocyte inclusions such as Howell-Jolly bodies, Heinz bodies and parasites [12]. On the other hand, manual counting is very subjective because of morphologic continuum between aggregate and punctate reticulocytes, which can lead to considerable variations and inaccuracies in the reported results [12]. In this paper we show that deep learning, given sufficient data, can provide advantages over both manual and automated methods. Computer vision, compared with the manual method, is more objective and faster at counting cells from images, and compared to automated methods, is more resilient to the presence of artifacts or anomalies. In addition, this method is less expensive than automated methods.

## 2. Using Deep Learning to Detect Reticulocytes

Our goal was to train a neural network to determine the ratio of aggregate reticulocytes to punctate reticulocytes and erythrocytes in a dataset of images of cat blood smears. We did this by training a convolutional neural network to perform 2D object detection on camera images. This means that for a given image, the network is supposed to output the locations of all objects of interest within the image in the form of a rectangular bounding box around each detected object indicating the location of the object, a classification label indicating what object the bounding box contains, and a confidence score indicating how certain the model is about the respective prediction. The following sections describe our dataset, the model we trained and how we trained it, and why we chose to formulate this task as an object detection problem.

## 2. 1 Dataset

The dataset used in our work was created using equipment that is easily accessible to veterinarians: a standard laboratory microscope and two types of cameras: a basic microscope camera and a smartphone camera. 1046 images were collected (553 images were taken with a "Bresser" microscope camera and 493 images with a "Samsung Galaxy S6" smartphone camera) from archived peripheral blood smears of two non-anemic male cats. Although the cats were non-anemic, these blood smears were chosen because of their high Heinz body content. As stated earlier, Heinz bodies are sometimes misclassified as aggregate reticulocytes, so it was important to have a higher proportion of Heinz bodies in order for the model to learn to distinguish the two cell types. The blood smears had been stained with new methylene blue (NMB) dye using the standard procedure. All images were taken on only one microscope, but the amount of light and the iris diaphragm aperture were randomly changed in order to create a dataset that would allow the model to generalize to a wider range of optical conditions. The images were scaled using Lanczos resampling or cropped to spatial dimensions of 300x300 pixels and then labeled using the labelImg annotation tool [13].

The resulting dataset is naturally unbalanced - non-anemic cats have less than 1% aggregate reticulocytes. Class imbalance is a common problem in deep learning that can lead to a bias towards predicting the majority class [14]. In order to mitigate this problem, the cells were divided into three categories: aggregate reticulocytes, punctate reticulocytes and erythrocytes. Also, the dominance of the erythrocyte class was reduced by manually removing a fraction of the erythrocytes from the images until the ratio of the three cell categories was approximately 1:1:2 (963 aggregate reticulocytes, 966 punctate reticulocytes, 1892 erythrocytes).

The images were randomly divided into two sets. 800 images (76.5% of the overall dataset) were put in a training set and used to fine-tune a pre-trained model described below. In order to be able to measure the training success, 246 images (23.5% of the overall dataset) were set aside to form a so-called validation dataset. The point of the validation dataset is to measure the model's performance on images that it has not "seen" during the training. However, since decisions on hyper parameter values are made based on the model's performance on the validation dataset, the final model's performance on the validation dataset will no longer be entirely representative of its real-world performance.

Therefore, for the final evaluation of the model, two additional test datasets were created and labeled using the same criteria. One dataset was created using a microscope camera, the other using a smartphone camera. Both datasets contain 80 images. Also, both datasets contain the same number of aggregate reticulocytes, namely 78, and, for easier comparison of the results, were created on the same blood smear. These datasets differ from the dataset used during training in the sense that they haven't been artificially balanced and they have a much higher cell density, thus being more representative of real-world conditions.

## 2.2 Data Augmentation

Since our dataset is very small, we used common data augmentation methods to create more diverse training input. Specifically, each input image was randomly flipped horizontally with probability 0.5, translated by up to 50 pixels in both spatial dimensions with probability 0.5, scaled by a factor of between 0.5 and 1.5 with probability 0.5, and had its brightness changed between 0.5 and 2.0 of its original brightness (in the HSV color space) with probability 0.5. These random transformations were performed ad-hoc during the training. Since the objects in our images are relatively uniform in size, the random scaling is particularly important to make the model robust to the scale of the objects.

## 2.3 Model

The model we chose to solve the object detection problem described above is the Single Shot MultiBox Detector (SSD) architecture introduced by Wei Liu et al. [15]. Specifically, we used a Keras reimplementation of SSD[3] and we chose an SSD300 that had been pre-trained on the Microsoft COCO dataset as our as our starting point for the training.

Since the model was pre-trained on the MS COCO dataset and thus predicts 80 different classes, the weight tensors of the classification layers were randomly sub-sampled to predict only 3 classes. The model was fine-tuned for 40 epochs over the dataset using an Adam optimizer [16] with an initial learning rate of 0.001 and a learning rate decay of 0.0005. As we report below, and unsurprisingly, this model achieves very encouraging results on the task at hand.

The question that naturally arises in this context is why it is even useful or necessary to formulate this particular task as an object detection problem. After all, we don't actually care where the individual blood cells are located within a given image, we only need the model to tell us how many of each type are there. We tried the following naive approach as an experiment: We took our trained SSD300, removed all predictor layers, and used the conv9_2 layer with three output channels to try to directly predict a number for each cell type. The assumption was that as long as the number of objects of each class is within a certain fixed range in all training images, the model should be able to learn to predict object counts directly, or at least some numbers that are proportional to the actual object counts. Unfortunately, this approach failed - the model learned nothing. The question of why this approach fails remains to be examined in detail.

## 3. Results

In a test set created with a microscope camera our model correctly classified 77 cells as aggregate reticulocytes, misclassifying only one reticulocyte, and thus achieving an accuracy of 98.7%. At the same time, in a test set created with a smartphone camera, the model underestimated number of reticulocytes, predicting only 69 of 78 (88.5%) the total reticulocytes.

In order to measure the model against human performance, we compared the model's accuracy with the results achieved by two veterinarians on the same datasets manually. In comparison with the first dataset, where the misclassified aggregate reticulocyte was classified as punctate reticulocyte not only by the model, but also by both veterinarians, in the second dataset the model failed to correctly classify six cells that were clearly recognizable as aggregate reticulocytes and were classified correctly by both veterinarians.

---

[3] Code is available at: https://github.com/pierluigiferrari/ssd_keras

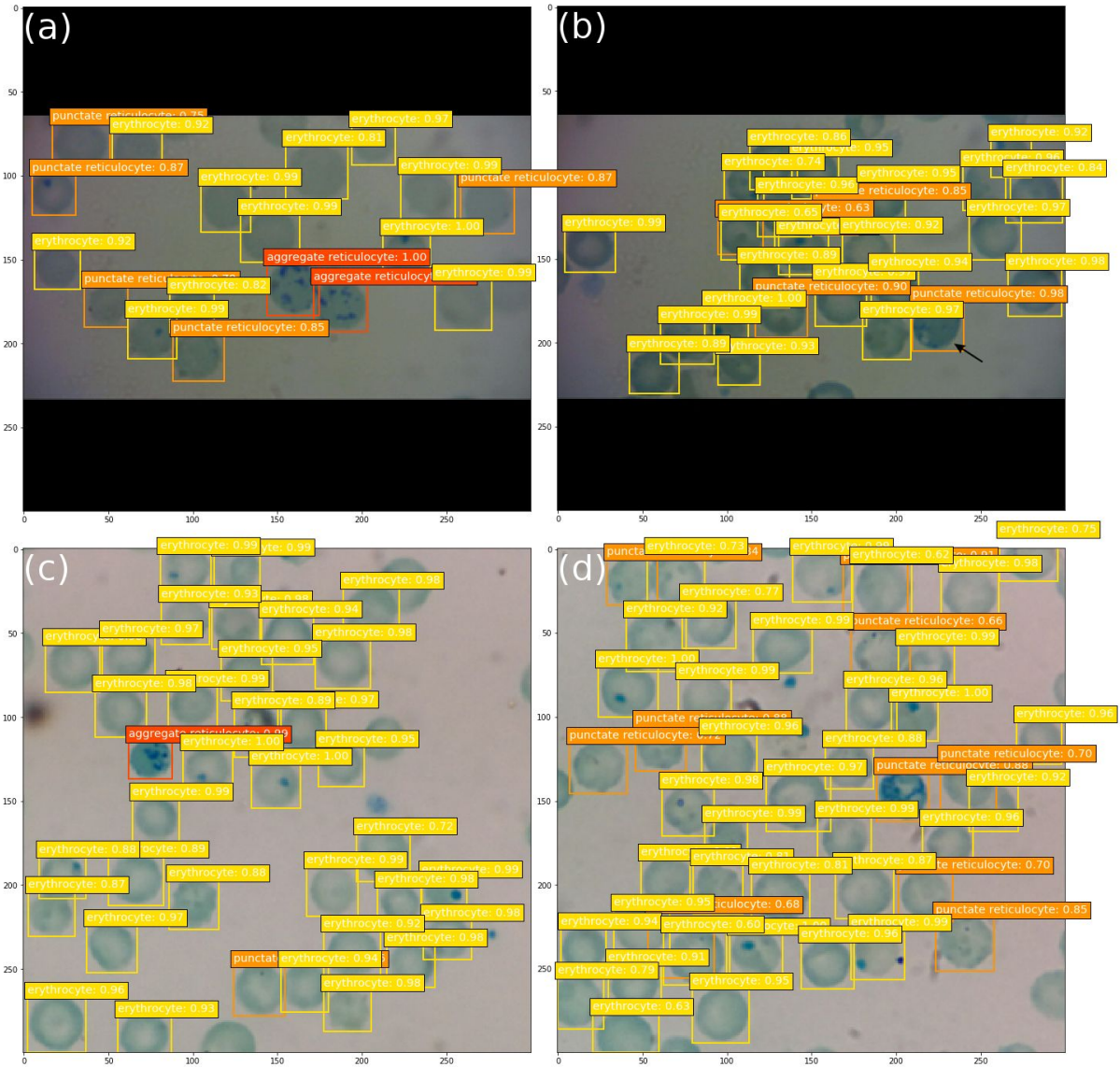

Fig. 2. Visualization of model predictions on four images from our test sets. Images (a) and (b) were taken with a microscope camera, images (c) and (d) with a smartphone camera. In image (b) the cell with the arrow was misclassified as a punctate reticulocyte both by the model and by the veterinarians. In image (d) model failed to correctly classify a cell that was clearly recognizable as aggregate reticulocytes and was classified correctly by both veterinarians.

The manual counting results of the two veterinarians were quite inconsistent, which was expected due to the subjectivity of manual counting. Compared with the model's results, they underestimated the reticulocyte count in both datasets and, notably, counted a significantly smaller number of reticulocytes in the dataset created with the smartphone camera (69 and 64 reticulocytes) than in the dataset created with the microscope camera (70 and 76 reticulocytes), despite the fact that both datasets have the same number of aggregate reticulocytes and were created on the same blood smear. Also, one veterinarian counted five reticulocytes more than the other in the first dataset and six reticulocytes less than the other in the second dataset, implying a variance that you certainly wouldn't want to accept from a method used in practice. This underscores the need for deep learning methods to get better in order to replace manual counting.

On the dataset created with the microscope camera, the results achieved by the model were, when taking into account both reticulocyte and erythrocyte counts, more closely correlated with our own manual results than those made by the veterinarians. On this dataset, our model calculated a reticulocyte percentage of 6.91%, which is only 0.13% less than our calculation (7.04%). The veterinarians, on the other hand, calculated reticulocyte percentages of 5.82% and 6.37%, respectively.

On the dataset created with the smartphone camera, the model fell short of the expected accuracy, predicting a significantly smaller reticulocyte count resulting in a reticulocyte percentage of 3.36%, which is 0.56% less than our result of 3.92%. The veterinarians reported results of 3.44% and 3.15%, respectively.

## 4. Discussion and Conclusion

In this paper we successfully implemented a convolutional neural network to determine the reticulocyte percentage from stained cat blood images. Using a pre-trained SSD300 model and fine-tuning it on only 800 labeled images, our model accurately classified 98.7% aggregate reticulocytes in images taken with a microscope camera and we can conclude that the training was successful. Furthemore, even when the model fell short of the expected results, as in the images taken with the smartphone camera, the predicted reticulocyte percentage was still within the error margin obtained by the veterinarians. Images taken with a smartphone camera are generally less uniform than those taken with a microscope camera, which makes the task a bit more difficult for the model. However, being a software-based solution, this method promises to get better over time as more images are collected and classified, increasing the dataset size, and better learning algorithms are developed.

We hope that this results will encourage others in the field to apply deep learning-based medical imaging to a number of problems in veterinary medicine. In some regards, veterinary medicine can be compared with human medicine in third world countries. Many basic laboratory tests are still too expensive for pet owners or are simply not feasible due to a lack of automated methods and the shear number of different species. For example, in birds and reptiles, white blood cells (WBC) count has to be done manually. This is because, unlike mammalian erythrocytes, erythrocytes in birds and reptiles have a nucleus. However, due to the number of different species and their morphological differences in WBCs, this test can only be done by specialists in this field and is therefore rarely used in practice. Deep learning can provide a solution to this and similar problems and bring much cheaper and more accurate diagnoses to veterinary medicine. In doing so, we believe that this technology can enable veterinary medicine to finally catch up with human medicine

For this to happen, though, this technology has to be accessible to everyone in this field. Often, the main limiting factor for implementing new technologies and methods in veterinary medicine is not the lack of knowledge, but the lack of resources. So, to illustrate that deep learning can be implemented by anyone in this field, we deliberately used equipment that is already accessible to most veterinarians: a standard laboratory microscope and a smartphone camera or very basic microscope camera (we should also note that the code and all the libraries used are open source). Also, deep neural networks for object detection are universally applicable to almost any domain and the model architecture we used for recognizing reticulocytes can also be used for any other object detection task given an appropriate dataset. The use of deep learning to perform a wide variety of tasks related to medical imaging only requires suitable data to train readily available models, and this is why veterinarians needs to be involved: only they have the medical expertise needed to create this data.